\documentclass[conference]{Definitions/IEEEtran}

\def\highlight{1}

\IEEEoverridecommandlockouts
\usepackage{cite}
\usepackage{amsmath,amssymb,amsfonts}
\usepackage{algorithmic}
\usepackage{graphicx}
\usepackage{graphicx}
\usepackage{threeparttable}
\usepackage{float}
\usepackage{makecell}
\usepackage{textcomp}
\usepackage{xcolor}
\usepackage{orcidlink}
\usepackage{datetime}
\usdate
\usepackage{lipsum}  
\usepackage{booktabs}
\usepackage{array}
\usepackage{caption}
\usepackage{siunitx}
\providecommand{\qty}[2]{\SI{#1}{#2}}
\providecommand{\unit}[1]{\si{#1}}
\usepackage{newtxmath}
\usepackage{altsubsup}
\SetAltSubscriptCommand{\mathrm} 

\DeclareSIUnit\mT{\milli\tesla}
\DeclareSIUnit\uT{\micro\tesla}
\DeclareSIUnit\nT{\nano\tesla}
\DeclareSIUnit\Nm{\newton\meter}
\DeclareSIUnit\mNm{\milli\newton\meter}
\DeclareSIUnit\uTNm{\uT\per\newton\per\meter}
\DeclareSIUnit\uTnoise{\uT\per\Hz\tothe{0.5}}
\DeclareSIUnit\pctFS{\percent\of{FS}}
\DeclareSIUnit\ppm{ppm}
\def\uT{\si{\uT}}
\def\Nm{\si{\Nm}}
\def\uTNm{\si{\uTNm}}

\def\mNm{\si{\mNm}}
\def\uTnoise{\si{\uTnoise}}
\def\uT_sqrthz{\si{\uTnoise}}

\def\mm{\si{\milli\meter}}
\def\cm{\si{\centi\meter}}

\def\mT{\si{\milli\tesla}}
\def\nT{\si{\nano\tesla}}

\newcommand\muted[1]{%
  \bgroup
  \hskip0pt\color{black!20!}%
  #1%
  \egroup
}
\usepackage{hyperref}
\hypersetup{ colorlinks, citecolor=teal, linkcolor=teal, urlcolor=teal}
\hypersetup{%
  hidelinks=true,
  bookmarksnumbered=true,%
  bookmarksopen=true}

\def\BibTeX{{\rm B\kern-.05em{\sc i\kern-.025em b}\kern-.08em
    T\kern-.1667em\lower.7ex\hbox{E}\kern-.125emX}}
\DeclareRobustCommand*{\IEEEauthorrefmark}[1]{%
  \raisebox{0pt}[0pt][0pt]{\textsuperscript{\footnotesize #1}}%
}
\usepackage{placeins}
\usepackage{dblfloatfix}
\usepackage{doi}
\usepackage{xcolor}
\ifdefined\highlight
  \newcommand\hl[1]{%
    \bgroup
    \hskip0pt\color{red!80!black}%
    #1%
    \egroup
  }
\else
  \newcommand\hl[1]{%
    #1%
  }
\fi

\newcommand\hlblue[1]{%
    \bgroup
    \hskip0pt\color{blue!80!black}%
    #1%
    \egroup
}

\renewcommand{\IEEEbibitemsep}{0pt plus 0.5pt}
\let\oldbibliography\bibliography
\renewcommand{\bibliography}[1]{%
  \begingroup
  \fontsize{7pt}{8pt}\selectfont
  \oldbibliography{#1}%
  \endgroup
}

\usepackage{tikz}\usetikzlibrary{calc}
\newcommand\version[1]{\tikz[overlay,remember picture]{\node at ($(current page.west)+(1.0,0)$)[rotate=90]
{\textcolor{gray!00}{\small v#1. Compiled on \today\ at \currenttime.}};}}

\usepackage{fancyhdr}
\fancypagestyle{topcopyright}{
  \fancyhf{}
  
  \fancyhead[C]{\fontsize{7.5pt}{8.5pt}\selectfont
    \textbf{Author's manuscript: version accepted for publication in 2026 IEEE SENSORS.}\\
    \copyright~2026 IEEE. Personal use of this material is permitted.
    Permission from IEEE must be obtained for all other uses, in any current or future media, including reprinting/republishing this material for advertising or promotional purposes, creating new collective works, for resale or redistribution to servers or lists, or reuse of any copyrighted component of this work in other works.
  }
}

\hypersetup{
    colorlinks,
    linkcolor={black!50!black},
    citecolor={black!50!black},
    urlcolor={black!50!black}
}

\begin{document}
\title{A Disk-Shaped Magnetoelastic Torque Sensor for Robotic Joints Using Permanent Magnetization} 
\newcommand{\orcidENG}{\orcidlink{0009-0004-2255-5952}}
\newcommand{\orcidBUR}{\orcidlink{0000-0003-4663-5071}}
\newcommand{\orcidLBA}{\orcidlink{0000-0000-0000-0000}}
\newcommand{\orcidNYZ}{\orcidlink{0000-0002-5903-1763}}
\newcommand{\orcidGCL}{\orcidlink{0000-0002-5140-7789}}
\newcommand{\orcidLSA}{\orcidlink{0009-0006-1574-8128}}
\newcommand{\orcidFCO}{\orcidlink{0000-0003-4415-5226}}
\newcommand{\orcidGBG}{\orcidlink{0000-0001-9972-0232}}
\newcommand{\orcidEKL}{\orcidlink{0009-0000-0512-8597}}

\author{\IEEEauthorblockN{
Bruno Brajon\IEEEauthorrefmark{1}\orcidBUR{},
Enrico Gasparin\IEEEauthorrefmark{1}\orcidENG{},
Nicole Yazigy\IEEEauthorrefmark{1}\orcidNYZ{},
Lisa Salamin\IEEEauthorrefmark{2}\orcidLSA{}\\
Florian Copt\IEEEauthorrefmark{2}\orcidFCO{},
Guzmán Borque Gallego\IEEEauthorrefmark{2}\orcidGBG{},
Elias Klauser\IEEEauthorrefmark{2}\orcidEKL{}, and
Gaël Close\IEEEauthorrefmark{1,*}\orcidGCL{}}
\IEEEauthorblockA{\IEEEauthorrefmark{1}Melexis Technologies SA, Bevaix, Switzerland}
\IEEEauthorblockA{\IEEEauthorrefmark{2}CSEM SA, Neuchâtel, Switzerland}\\
\IEEEauthorblockA{\IEEEauthorrefmark{*}Corresponding author: gcl@melexis.com}
\thanks{This project received the support of Innosuisse (project n. 123.079 IP-ENG).}
}

\maketitle
\thispagestyle{topcopyright}

\begin{abstract} 
Direct torque sensing is a growing need in the robotics community
to enable precise control and interactions where torque estimation from motor current
is not sufficient. This paper presents a novel disk-shaped magnetoelastic torque sensor
with a compact axial envelope of about \qty{1}{\cm}, suitable for integration in robotic joints.
A four-magnetometer architecture is used to measure the field modulated by the stress
affecting a narrow magnetized region while 
rejecting the effects of parasitic cross forces.
A custom-designed magnetic shield enhances the torque sensitivity
while reducing the external stray fields by a factor 7x. 
The device measures the torque with an accuracy of \qty{1.34}{\pctFS}
relative to the \qty{50}{\Nm} full scale (FS).
The paper details the development of the sensor
through the mechanical design, the magnetization procedure,
and the experimental validation. The results demonstrate the potential
of the proposed sensor for robotic applications.

\end{abstract}
\section{Introduction}
\label{sec:intro}

As robots are leaving their cages and entering environments
populated by humans, there is a growing need for granting safe human-robot interactions
\cite{SHAH2025105704}. 
Torque sensors integrated into the robotic joints
allow real-time feedback of the environmental forces
to avoid potential hazards \cite{ZAFAR2024102769}. 
Safety standards for collaborative robots now impose strict contact-force limits
that require real-time closed-loop torque feedback with the use of torque sensors
\cite{iso15066_2016}. 

Sensorless approaches, 
where the torque is estimated from the motor's current and a model,
have been proposed to avoid the cost and complexity of physical sensors.
These are fundamentally limited by uncertainties in friction, backlash, 
and gear-train parameters, 
all of which degrade with temperature, speed, and mechanical wear 
\cite{deluca_collision_2006, haddadin_robot_2017}.
Hence, torque estimation becomes unreliable 
especially in high-gear-ratio transmissions
where gear friction and other energy losses make the motor side blind
to the actual load torque \cite{wahrburg_motor_2018}.

Today, most of the torque sensors used in robotics
are based on strain gauges.
They set the accuracy standard for joint torque sensing. 
They require the sensing elements to be glued on the holding structure, 
making them potentially susceptible
to long-term drifts \cite{Ogushi2023}. 

By contrast, in the automotive industry for steering applications,
contactless magnetic or inductive position sensors based on torsion bar
are used for their robustness and low cost \cite{angleviel_development_2006}.
Such deformable bars are less suitable for most robotic applications,
which require high mechanical stiffness. 

Magnetoelastic torque sensors allow contactless sensing
and high torsional stiffness, a unique desirable combination. 
The technology is widely deployed
in e-bike applications in high-volume production,
given its proven robustness, long-term stability
and reduced manufacturing cost \cite{julius_2021}.
The working principle relies 
on the variation of the magnetic permeability in steel due to stress.
Two sensing principles can be identified.
In the active readout, 
an AC probing field is actively applied to the shaft via a coil,
and the induced magnetic response is measured inductively \cite{ZHANG2025100229}.
In the passive readout, which is used in this work, 
a portion of a cylindrical steel shaft is subjected
to a circular permanent magnetization instead of an AC probing field.
Under the influence of torque, the circular magnetization tilts 
towards the direction of principal stress, the \ang{45}-oblique,
and acquires an axial component as a result.
Hall or fluxgate magnetometers can be used to measure the resulting field
\cite{patent_lee_magnetoelastic_2013, patent_ay_magnetic_2010}.
In order to achieve stray-field immunity,
these sensors are often arranged in a differential fashion along the shaft,
often requiring an axial footprint of several centimeters
\cite{garshelis_development_1997, muro_magnetostrictive-ring_2014}.

\begin{figure}[t!]
    \centering
    \includegraphics[width=\columnwidth]{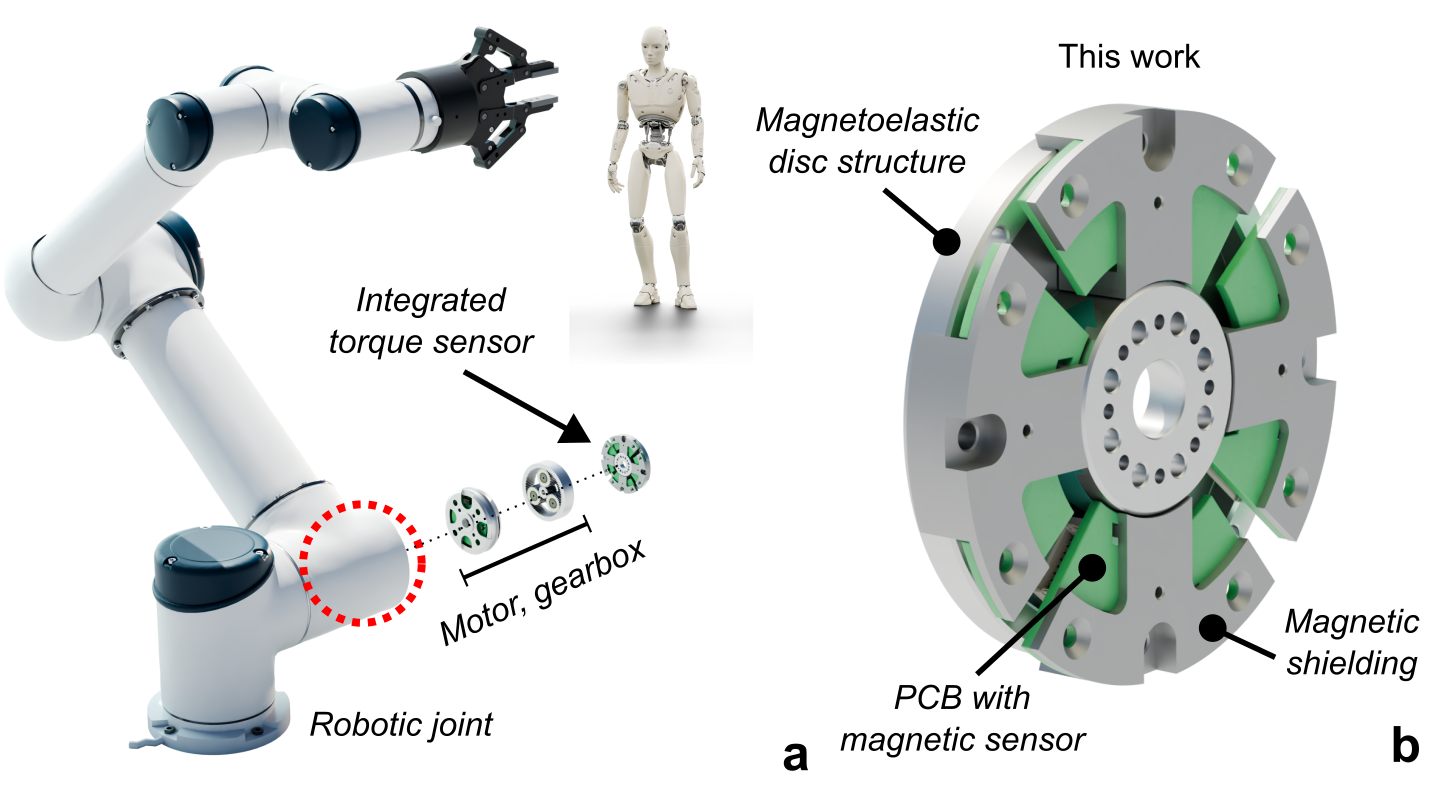}
    \caption{Conceptual diagrams.
    \textbf{(a)} Example of torque sensing in robotic joint.
    \textbf{(b)} Proposed sensor concept.}
    \label{fig:rd116_f1_concept} 
\end{figure}

In stark contrast to the state-of-the-art magnetoelastic sensors in e-bikes, 
this work demonstrates a disk-shaped geometry (see \autoref{fig:rd116_f1_concept})
that is suitable for integration in robotic joints.
To the best of our knowledge, this work is the first 
to exploit the passive magnetoelastic effect
in such a compact form.
\section{Sensor Design and Implementation}
\label{sec:methods}

In prior work, we demonstrated a Hall-based magnetoelastic sensor 
for e-bikes \cite{gasparin_magnetoelastic_2024},
with one or more magnetized regions
along the classical long shaft (\qty{140}{\mm}) 
found in e-bikes bottom brackets (between the pedals).
To adapt this sensor to a disk-shaped structure,
the transducer was redesigned as a \qty{1}{\mm} thin wall cylinder with 
\qty{28}{\mm} inner diameter and \qty{30}{\mm} outer diameter.
The thin-walled cylinder is connected 
to flanges on both sides.
It is a monolithic element made of X30Cr13,
a martensitic stainless steel, in the category of FeNi alloys
with expected saturation magnetostriction in the order of
$\lambda_\mathrm{s} \approx \qty{10}{\ppm}$ \cite{10818450}.
The magnetization is applied to the thin wall cylinder which will act as transducer.
The geometry allows high stress concentration in this region
to enhance the variation of the output magnetic field due to torque.
In fact, the rotation of the magnetization vector
depends on the stress $\sigma$, as in the equation
derived from the magnetic energy \cite{cullity_introduction_2009}

\begin{equation}
\theta_[M] = \frac{1}{2} \arctan\left( \frac{3\lambda_\mathrm{s}\sigma}{2K_[u]} \right),
\end{equation}

where $K_[u]$ is the magnetic anisotropy constant. 

The disk is designed for a full-scale torque of \qty{50}{\Nm},
an overload capacity of \qty{150}{\Nm},
and parasitic forces up to \qty{500}{\newton}, 
resulting in a maximum calculated stress of \qty{250}{\mega\pascal}.
The mechanical stiffness of the joint is
\qty{125}{\kilo\newton\meter\per\radian},
sufficiently high for most robotic applications.
The smaller inner diameter of the part is \qty{10}{\mm} to allow cable routing.

Four magnetometers (with outputs $U_1$, ..., $U_4$),
disposed at \ang{90} intervals around the circumference 
measure the radial component of the magnetic field $B_\mathrm{r}$ appearing 
when torque is applied.
The signals' average $\overline{U}$ is used to estimate the torque. 

Simulations show that this configuration is robust 
against parasitic axial and radial forces acting on the structure.
Those produce a magnetic response whose amplitude
follows a sine wave over the angle on the circumference,
while the nominal torque signal is a constant. 
Thus these contributions on $\overline{U}$ are balanced out. 

\begin{figure}[htbp]
  \centering
  \includegraphics[width=\columnwidth]{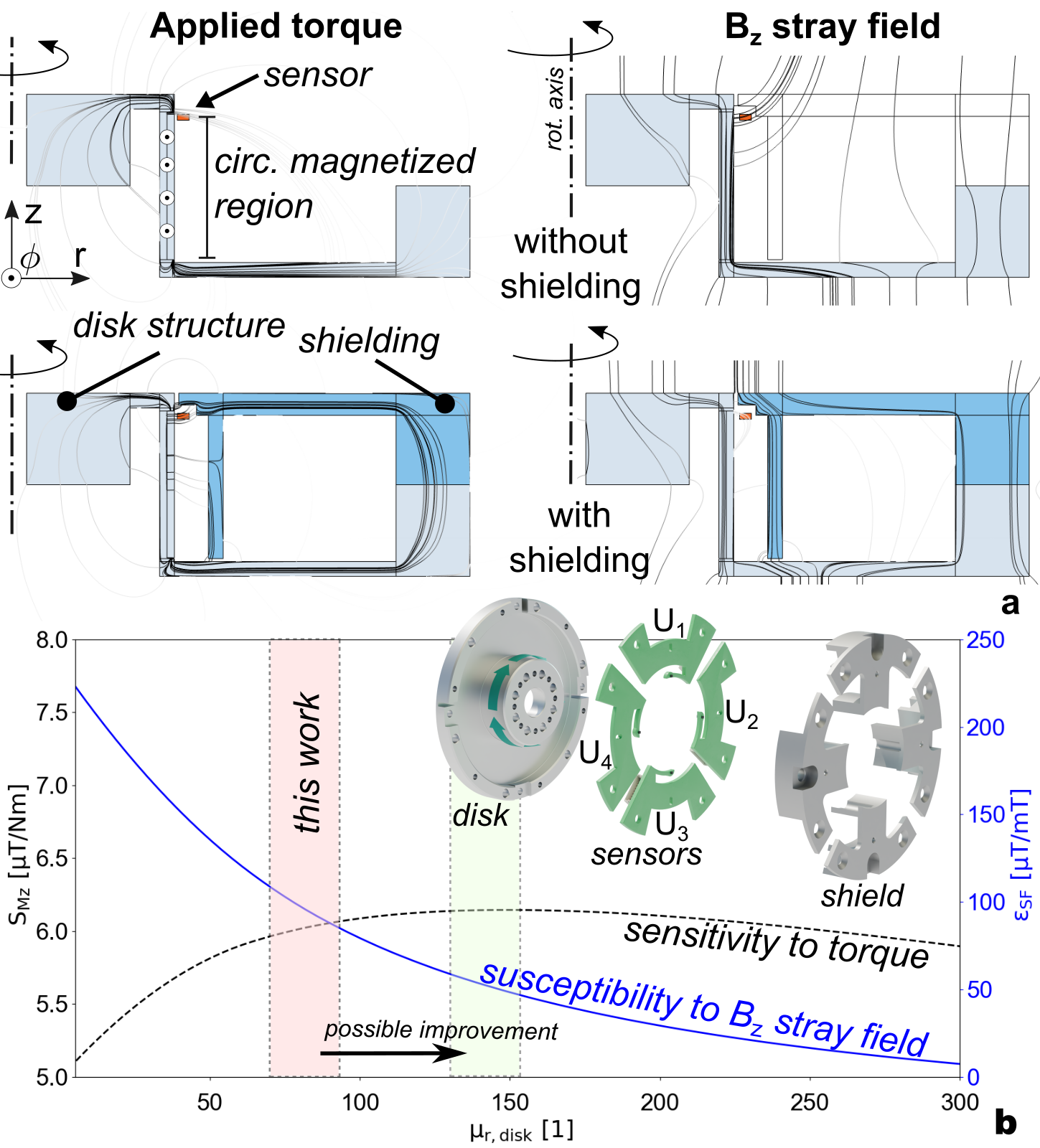}
  \caption{Simulation results \textbf{(a)}  showing the impact of the shielding on the
  sensitivity to applied torque or axial stray fields
  and \textbf{(b)}  effects of the variation of the 
  relative magnetic permeability of the disk structure.}
  \label{fig:simulation_concept} 
\end{figure}

\begin{figure}[htbp]
  \centering
  \includegraphics[width=\columnwidth]{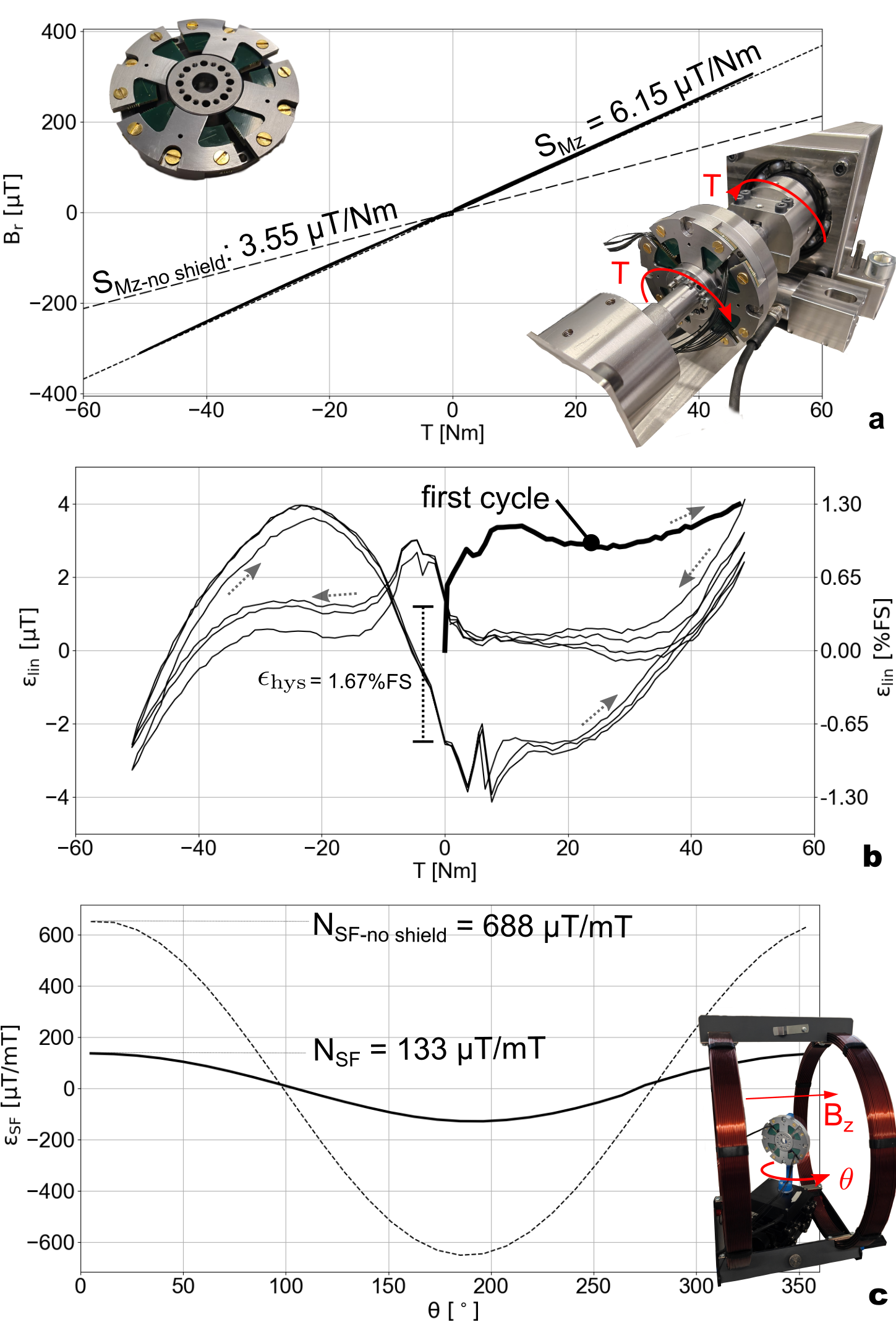}
   \caption{
    Experimental results: 
\textbf{(a)} Sensitivity to applied torque,
   \textbf{(b)} linearity error over the full scale,
   \textbf{(c)} stray field error measured at different amplitudes.
   }
  \label{fig:results_panel} 
\end{figure}

The magnetometers are covered with magnetic shielding structures
designed to reduce the effects of external magnetic fields. 
This also contributes to increasing the sensitivity to torque,
$S_{Mz}$, expressed in \unit{\uTNm},
by almost a factor of 2 (\autoref{fig:results_panel}a). 
The shielding is made of AISI 1010, in annealed state,
a common low-carbon steel with initial relative magnetic permeability
$\upmu_{r,\mathrm{shield}} > 500$ \cite{tortolero_2021}. 

Due to the structure's symmetry we can consider the stray fields on two axes:
radial and axial. Radial stray-field rejection
does not depend on the shielding but rather on the signals' averaging:
the disturbance appears with opposite signs on 
two diametrically opposed magnetometers. 
Axial stray-field rejection relies 
on the relative positioning of the magnetometer
and the shielding and their relative permeability, 
which creates a zero-stray-field point
(\autoref{fig:simulation_concept}(a)).
Simulations capture the effect of permeability variation
of the magnetized element on the sensitivity $S_{Mz}$
and the residual error due to axial stray fields
$\epsilon_{\mathrm{SF}}$ (\autoref{fig:simulation_concept}(b)).
After comparison to the experimental results, 
we estimate the structure permeability
to be $\upmu_{r,\mathrm{disk}} \lessapprox 100$, 
which is reasonable given the hard magnetic properties of the material,
necessary for the permanent magnetization. 

The magnetometers (MEMSIC MMC5603NJ)
are 3-axis AMR with a noise floor of \qty{0.5}{\uTnoise}
and a full-scale range of $\pm$\qty{3}{\mT} \cite{mmc5603nj_datasheet_2022}.
The wide range provides headroom to avoid saturation
in presence of stray fields.
\section{Experimental Results}
\label{sec:results}

\begin{table*}[!b]
    \begin{minipage}[c]{0.73\linewidth}
        \caption{Comparison of torque sensors.}
        \vspace{-2pt}
        \label{table:comparison}
        \begin{center}
        {\setlength{\tabcolsep}{3pt}\small
        \begin{tabular}{>{\raggedright\arraybackslash}p{2.8cm} >{\raggedright\arraybackslash}p{2.0cm} >{\raggedright\arraybackslash}p{2.0cm} >{\raggedright\arraybackslash}p{2.0cm} >{\raggedright\arraybackslash}p{2.0cm}}
            \toprule
            \textbf{ }
            & \textbf{This work}
            & \textbf{NCTE~\cite{ncte_sbbrt_datasheet}}
            & \textbf{AIDIN~\cite{aidin_atsb50_datasheet}}
            & \textbf{ForceN~\cite{forcen_dev1ja_datasheet}}
            \\ \midrule

            Application
            & Robotics
            & E-bikes
            & Robotics
            & Robotics
            \\

            Technology
            & Magnetoelastic
            & Magnetoelastic
            & Capacitive
            & Strain Gauge
            \\

            Size (D$\,{\times}\,$H) [mm]
            & \O{}70$\times$12.4
            & \O{}17$\times$140
            & \O{}84$\times$14
            & \O{}61$\times$9.5
            \\

            Full scale [\unit{\Nm}]
            & 50
            & 200
            & 50
            & 45
            \\

            Resolution$^*$ [\unit{\mNm}]
            & 41.4
            & 240
            & 7.07
            & 2.65
            \\

            NFR$^*$ [bits]
            & 8.51
            & 8
            & 11
            & 12.3
            \\

            Accuracy [\unit{\pctFS}]
            & $\pm$1.34
            & $\pm$2.5
            & --
            & $\pm$1
            \\

            \bottomrule
        \end{tabular}
        \par\vspace{2pt}
        \raggedright\footnotesize $^*$Calculated over a normalized sampling rate of \qty{500}{Hz}.
        }
        \end{center}
    \end{minipage}%
    \hfill
    \begin{minipage}[c]{0.30\linewidth}
        \centering
        \includegraphics[width=\linewidth]{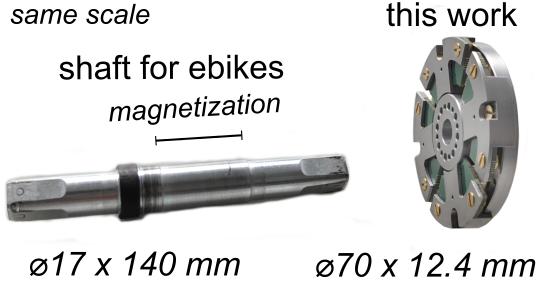}
        \captionof{figure}{Size comparison.}
        \label{fig:concept_comparison}

    \end{minipage}
\end{table*}

\subsection{Sensitivity}
The sensitivity has been measured using a static torque bench 
with a reference torque sensor. 
A motor provides a controlled load ranging from \qty{-50}{\Nm} to \qty{50}{\Nm} 
while the magnetic field is recorded 
at a sampling rate of \qty{500}{Hz} by the four magnetometers
mounted at a \qty{0.6}{\mm} air gap from the magnetized surface.
The signal average $\overline{U}$ reveals
a RMS noise of \qty{255}{\nT} 
and a sensitivity to torque $S_{Mz} = \qty{6.15}{\uTNm}$,
(\autoref{fig:results_panel}a).
In mechanical units, it is equivalent to 
a RMS noise of $\tau_[n]=\qty{41.4}{\milli\Nm}$,
and a Noise Free Resolution 
$\mathrm{NFR}=\log_2(2\tau_[FS])/(6.6\tau_[n])$ 
of 8.51 bits for the full scale  $\tau_[FS] = \qty{50}{\Nm}$.

\subsection{Hysteresis and Linearity Errors}
The linearity and hysteresis errors have been calculated 
as the residuals of four bidirectional torque sweeps 
relative to the best-fit straight line. 
The linearity error $\epsilon_[lin]$ is calculated 
as the largest deviation of the residuals,
resulting in $\pm \qty{1.34}{\pctFS}$.
The hysteresis error $\epsilon_[hys]$ is calculated
as the variation of $\overline{U}$ at zero load across all cycles, 
resulting in $\pm \qty{0.833}{\pctFS}$, 
(\autoref{fig:results_panel}b).

\subsection{Stray-Field Immunity}
The susceptibility to stray fields has been quantified by applying a
uniform static magnetic field of up to $\qty{1}{\mT}$ with a Helmholtz coil arrangement.
The torque sensor module
is then rotated by \ang{360} over its radial axis.
The worst case scenario is obtained when the torque sensor
is aligned with the coils, at \ang{0}
and \ang{180}, (\autoref{fig:results_panel}c).
The susceptibility to stray field is reduced from
$S_[SF, no-shield] = \qty{688}{\uT\per\mT}$ without shielding, 
to $S_[SF] = \qty{133}{\uT\per\mT}$ in the proposed configuration.
Radial stray fields have a negligible influence
at \ang{90} and \ang{270}.
These are easily distinguishable from the torque-induced fields as they have opposite signs
and are readily averaged out.

\setlength{\parskip}{0cm}
\section{Conclusions} 
\label{sec:conclusions}
In this work, we presented a proof of concept
of a novel disk-shaped magnetoelastic torque sensor with integrated shielding,
suitable for integration into robotic joints.
The results of this work
are summarized in \autoref{table:comparison} 
and compared to the specifications of commercial torque sensors.

In terms of NFR and linearity error,
we obtained performance marginally exceeding that of an established magnetoelastic torque sensor
used in e-bikes. Both are based on a similar monolithic design
with permanent magnetization and magnetic sensors, 
but with radically different form factors
(Fig. \ref{fig:concept_comparison}). 
The shielding component was added to the sensing chain
with the double benefit of increasing the sensitivity
and providing partial rejection of stray fields.
We speculate that this can be improved at least by a factor 2, 
through proper selection of the processing steps.
There remains a performance gap with respect to capacitive and strain-gauge-based transducers.
Both offer higher resolution and accuracy.

While the current error budget provides a foundational assessment,
it remains partial.
Additional experiments are planned to quantify 
the susceptibility to
parasitic bending moments and forces,
as well as temperature fluctuations.

Overall, this work opens the door to the adoption of
magnetoelastic torque sensors in robotic joints.
The technology is proven in e-bikes in a safety-critical function
(providing controlled rider assistance) in high volumes
in outdoor environments at the right price point.
It has the potential for further improvements
to rival established strain gauge and capacitive torque sensors
in a similar disk-shaped form factor.

\FloatBarrier
\clearpage
\version{1.1.0}
\IEEEtriggeratref{10} 
\bibliographystyle{IEEEtran}
\bibliography{biblio}
\end{document}